\documentclass[letterpaper, 10 pt, conference]{ieeeconf}  

\IEEEoverridecommandlockouts                              

\usepackage{graphicx}
\usepackage{times}
\usepackage{epsfig}
\usepackage{amsmath}
\usepackage{amssymb}
\usepackage{verbatim}
\usepackage{multirow}
\usepackage{booktabs}
\usepackage{xcolor}
\usepackage{color}

\title{\LARGE \bf
Target Driven Instance Detection
}

\author{Phil Ammirato$^{1}$, Cheng-Yang Fu$^{1}$, Mykhailo Shvets$^{1}$, Jana Ko\v{s}eck\'{a}$^{2}$, Alexander C. Berg$^{1}$
\thanks{$^{1}$University of North Carolina at Chapel Hill, Computer Science
{\tt\small [ammirato, cyfu, mshevts, aberg]@cs.unc.edu}}%
\thanks{$^{2}$ George Mason University, Computer Science
        {\tt\small kosecka@gmu.edu}}%
}

\begin{document}

\maketitle
\thispagestyle{empty}
\pagestyle{empty}

\begin{abstract}
 While state-of-the-art general object detectors are getting better and better, there are not many systems specifically designed to take advantage of the instance detection problem. For many applications, such as household robotics, a system may need to recognize a few very specific instances at a time. Speed can be critical in these applications, as can the need to recognize previously unseen instances.  We introduce a Target Driven Instance Detector (TDID), a novel architecture for instance detection. TDID not only improves performance on instances seen during training, with a fast runtime, but is also able to generalize to detect novel instances. We demonstrate superior performance compared to standard object detection models as well as more traditional recognition approaches based on hand-engineered features on modern, challenging datasets. 

\end{abstract}

\section{INTRODUCTION}

Object detection works!  Alas, this is not always true, and the specific version of object detection matters.  There have been massive improvements in the accuracy of {\em category-level} object detectors based on deep learning~\cite{fasterRCNN,liu15ssd}.  These require many labeled training examples of bounding boxes for each category (e.g. mug) in question, use carefully crafted data augmentation approaches to fully leverage that training data, and can take days or longer to train.  This leaves out an important type of object detection problem where the goal is to detect a precise instance of an object category (e.g. my mug instead of a mug).  This setting applies to real world tasks including fetch and deliver in household environments, and robotic manipulation in industrial environments, where the objects in question are often specific instances and not general categories.  The instance task does not have the large intra-class variation of category-level detection, and sometimes only a small number of training examples for each instance is available.

How can the progress on category-level object detection be harnessed and applied to instance detection, taking advantage of the specificity of instances and overcoming the challenge of small numbers of training examples?  One approach is to take a small number of example images and artificially create a large number of detection training examples by artificially composing those examples into scenes~\cite{cutpaste2017,gmuSynthDetection}.  This still treats instance detection as a category detection problem, but expands a small number of clean images of an object instance into enough samples to train current category detectors. Another possible approach reduces the training necessary for new targets by preconditioning a network to be robust to varying views of objects. Recent work by ~\cite{instanceHeld} has shown good accuracy with such an approach, and that a deep-learning-based method for learning a classifier from single examples can be more accurate than a wide range of previous template matching approaches.

This paper presents a new approach that goes further than the two above by learning a detector that directly takes advantage of the uniqueness of instances, and that does not need to be retrained or fine tuned in order to detect a new target object.  This is done by learning an embedding that compares learned features of the target to learned features at each location in a scene image, and integrating this into a state-of-the-art detection framework.

\begin{figure}[t]
\begin{center}
   \includegraphics[width=0.8\linewidth]{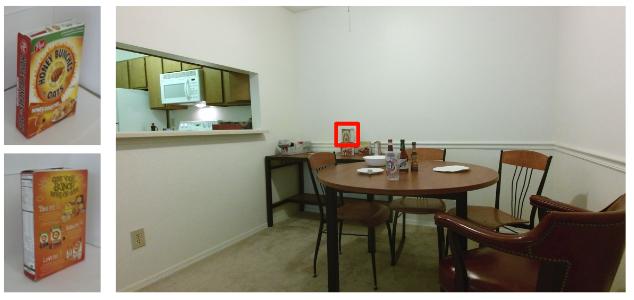}
\end{center}
   \caption{Example of target images (front and back view of the object), and an input ``scene image'' that contains the target object in a different pose, partially occluded,  at small scale. The object's bounding box (in red) is included in the figure for reference.}
\label{fig:target_input_ex}
\end{figure}

The Target Driven Instance Detection problem is formulated as follows: given an input scene image $S$ and a number, $T$, of images of a target object, output a bounding box around the target object in $S$, or no box if the object is not present. See Figure \ref{fig:target_input_ex} for an example with $T=2$ target images with the correct output shown. 


The datasets~\cite{active-vision-dataset2017,gmuKitchenDataset,UWASH-OBJECTS} we use to evaluate our Target Driven Instance Detector (TDID) display a set of object instances in everyday home environments and exhibit real-world confounding factors such as scale variation, clutter and occlusion.  An important aspect of the Active Vision Dataset~\cite{active-vision-dataset2017} (used for test in many experiments) is that it was collected to sample views of household rooms from every position where a robot could navigate.  As a result, many objects are quite small in some views, perhaps when seen a
cross a room, and are partially occluded in many views.

The objects used in our experiments come from the BigBird and RGB-D Object datasets~\cite{BigBIRD,UWASH-OBJECTS}, and we note that part of the methods success stems from seeing similar objects in training.  This is the same in previous work to which we compare, and is reasonable to expect in real-world mobile manipulation applications, but it is important to make this clear.

\begin{figure}
\begin{center}
   \includegraphics[width=0.45\textwidth]{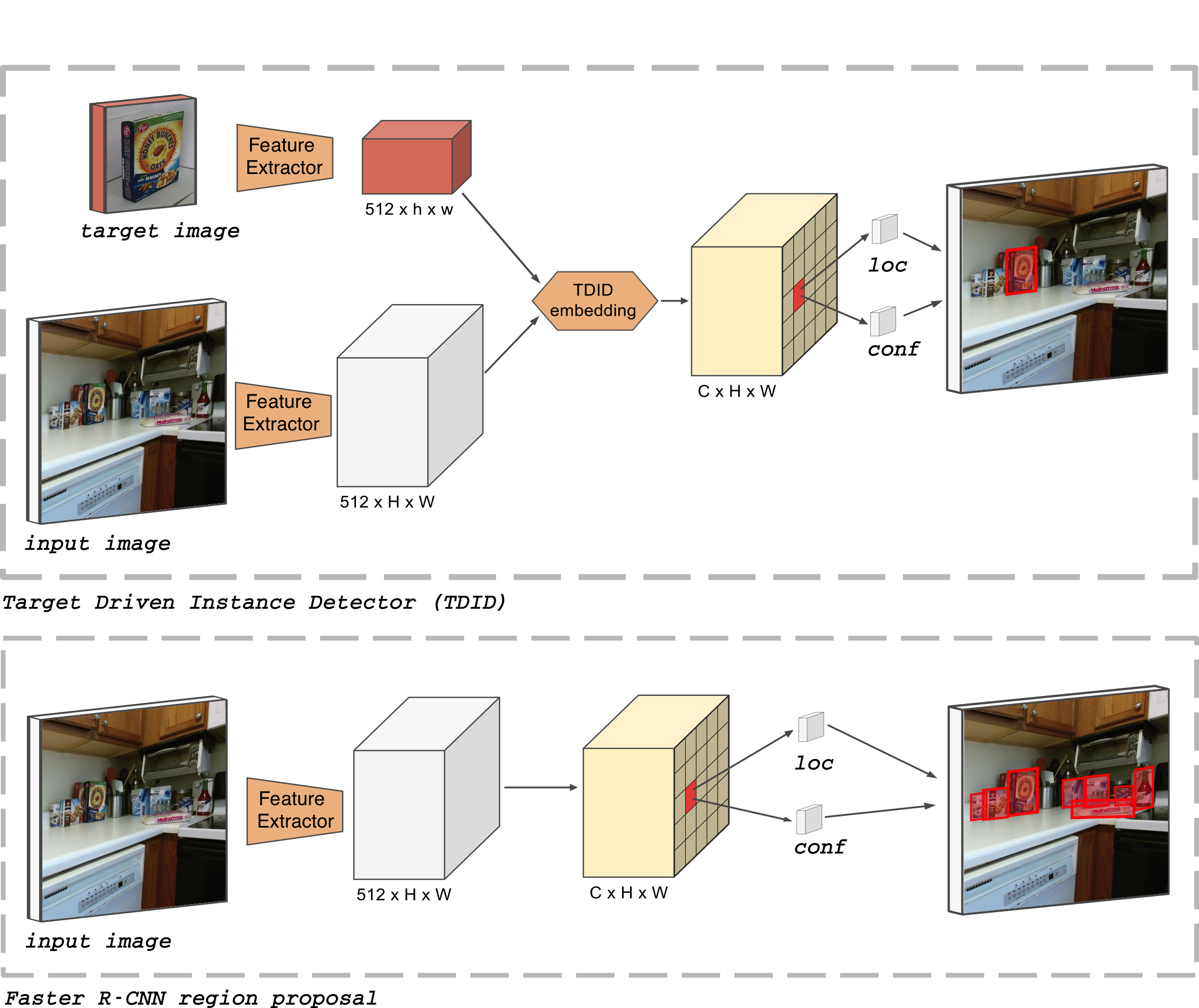}
\end{center}
    \caption{The proposed architecture. The bottom box represents the RPN stage of Faster R-CNN. The top box is our TDID model. We enrich the feature representation with a joint embedding for scene-target pair. TDID first extracts features from scene and target image (feature extractor weights are shared), then combines those with a novel TDID embedding module, and finally applies the detection prediction head. The detailed structure of the TDID embedding module is shown in Figure~\ref{fig:tdid_embedding} and described in Section~\ref{sec:method}.}
\label{fig:arch_full}
\end{figure}

We summarize our contributions as follows: \emph{(1)} A novel \emph{Target Driven Instance Detector} (TDID) model that easily transforms the current state-of-the-art general object detectors into instance detectors, as depicted in Figure~\ref{fig:arch_full}. \emph{(2)} Strong performance improvement on multiple challenging instance detection scenarios.  \emph{(3)} We compare TDID to previous work on one-shot training for instance classification, including more classic template matching, and show better accuracy. \emph{(4)} The ability to generalize detection to unseen instances on challenging datasets without any additional training or fine-tuning.

\section{Related work}


Traditional methods for object detection in cluttered scenes follow the sliding window based pipelines where efficient methods for feature computation and classifier evaluation were developed such as DPM~\cite{Felzenswalb_TPAMI10}. Examples of using these models in the table top setting  include~\cite{Lai_ICRA11,Song_ECCV12}.  Object detection and recognition systems that deal with textured household objects such as Collet et al.~\cite{collet_IJRR11} and Tang et al.~\cite{Tang_ICRA12} take advantage of the discriminative nature of local descriptors. A disadvantage of these local descriptors is that they usually perform poorly in the presence of non-textured objects. Some of these issues were tackled by~\cite{lineMOD_ACCV11} which used template based methods to deal with such texture-less objects.  Hand engineered features typically work well in table top settings that contain a relatively small number of objects at relatively large scale~\cite{zeng2018robotic}. The authors in~\cite{lepetit_CVPR15} introduce an effective approach to feature learning for simultaneous categorization and pose estimation for single objects on uniform backgrounds.

\textbf{General Object Detector}
State-of-the-art object-category detectors have been improved significantly over the last few years in both accuracy and speed. These detectors rely on a backbone architecture, such as VGG~\cite{vgg} or ResNet~\cite{resnet}, to extract features from the image, and then add a detection module on top of these features. Two-stage detectors, such as Faster R-CNN~\cite{fasterRCNN}, and R-FCN~\cite{dai16rfcn}, rely on an initial region proposal followed by a classification and location regression of the proposed regions. Recent single-stage detectors: YOLO~\cite{yolov1}, YOLOv2~\cite{yolo9000}, and SSD~\cite{liu15ssd} skip the feature pooling stage and show that fast inference speed can be achieved. Recent work has added top-down connections~\cite{lin2016fpn,dssd,tdm_arxiv16}, which can borrow rich semantic information from deeper layers and improve in accuracy for small objects, though usually at reduced speed.

\textbf{Instance Recognition}
Compared to object-category recognition, the specific instance recognition setting has less intra-class variation and, is often allowed a limited number of training examples. Much work has been done using hand-crafted features and template matching to identify object instances even since somewhat recent seminal work \cite{SIFT,fergus2003objectRecognition}.  More recently hand-crafted features have been replaced with learned ones in many recognition tasks \cite{Krizhevsky_imagenetclassification}. The instance recognition dataset BigBIRD \cite{BigBIRD}, which provides dense, individual scans of over 100 object instances on a turntable has enabled more research on instance recognition. \cite{instanceHeld} shows that pre-training on BigBIRD improves robustness to pose and improves classification performance over hand-crafted and template matching methods, even if only one image per instance is provided for training.

\textbf{Instance Detection} 
The recent release of larger scale instance detection datasets like the Active Vision Dataset(AVD)~\cite{active-vision-dataset2017} and GMU Kitchens~\cite{gmuKitchenDataset}, has enabled more work using deep learning for instance detection. The GMU Kitchen Dataset has 6,728 images across 9 scenes, and the initial release of AVD has 17,556 images across 9 scenes. Both datasets feature instances very similar to those in BigBIRD \cite{BigBIRD}, with GMU featuring 11 such instances and AVD 30. \cite{cutpaste2017,gmuSynthDetection} attack the problem of limited training examples by synthesizing new examples with different background images. In both of these works general object category detectors such as SSD~\cite{liu15ssd} or Faster R-CNN~\cite{fasterRCNN} are still used to solve the instance detection problem.


\textbf{Navigation} ~\cite{zhu2017target} address a related problem, exploring an environment to reach a target position. They also input both a target image (of the desired view) and an image from the current position, and learn an embedding to aid in navigation. It is not straightforward to adapt their method to the instance detection problem, however, as they aim to move so that the image at the current position matches the target exactly. The embedding is not designed to localize objects, which is necessary for detection. Furthermore, the network requires scene-specific layers, while most object detectors are expected to generalize to unseen environments. 

\textbf{Tracking}
Given an initial bounding box of an object, the tracking task is to localize the same object appearing in each subsequent video frame. Correlation is frequently used for estimating similarity of patches between frames \cite{Bertinetto2015Stapple}  \cite{henriques2015KCF}
\cite{danelljan2014DSST}. Recent deep learning methods, such as \cite{trackingYCNN}, uses a Siamese network to measure the similarity in tracking. \cite{trackingFCSiam} uses a correlation filter to transform the Siamese network to be fully convolutional. \cite{Held2016Tracking100FPS} combines the features of crops from previous and current frames to regress the location directly. \cite{trackingendtoendCorrFilter} interprets the correlation filter learner as a differentiable layer and enables learning deep features that are tightly coupled to the correlation filter. Strong priors on the object and background exist in tracking, namely that neither changes much frame to frame. These priors include scale, location, illumination, orientation and viewpoint. Our instance detection setting requires robustness to larger changes between target and scene.

\section{Method}\label{sec:method}
\subsection{Problem Formulation}
Instance detection requires a system to recognize and localize specific objects in novel images. Usually these  \emph{scene images} images contain many objects, some of which are instances to be recognized. Most object detectors work by training on a set of scene images and ground truth bounding boxes of objects, and test on novel scene images containing the same types of objects. General object detectors attempt to find all object instances in a scene image at once. 

Our Target Driven Instance Detector (TDID), takes as input not only a scene image, but also one or more \emph{target images}. These target images contain only the instance of interest, see Figure \ref{fig:target_input_ex}. TDID attempts to detect only this target instance in the scene image.  

\subsection{Network Architecture}
TDID is similar to the Region Proposal Network (RPN), the first stage of Faster-RCNN, but adds a target/scene joint embedding. Figure \ref{fig:arch_full} compares our architecture with that of the RPN. With this joint embedding we are able to outperform other detectors even without the second stage of the traditional Faster-RCNN pipeline. This results in an architecture that computes detection outputs in one shot, with speed close to other one-shot detectors, while achieving better accuracy on various instance detection tasks than both one and two stage detectors. 

The high-level view of our architecture is as follows: Extract features from the target and scene images using some shared feature extraction network, such as VGG-16\cite{vgg}. Next, pass both target and scene image feature maps through our joint embedding. Finally, a set of convolutions predict class scores and bounding box regression parameters for a set of default anchors boxes (see Figure \ref{fig:arch_full}) over the embedding feature map. In TDID there are only two classes: target object or background.

\begin{figure}[t]
\begin{center}
   \includegraphics[width=.85\linewidth]{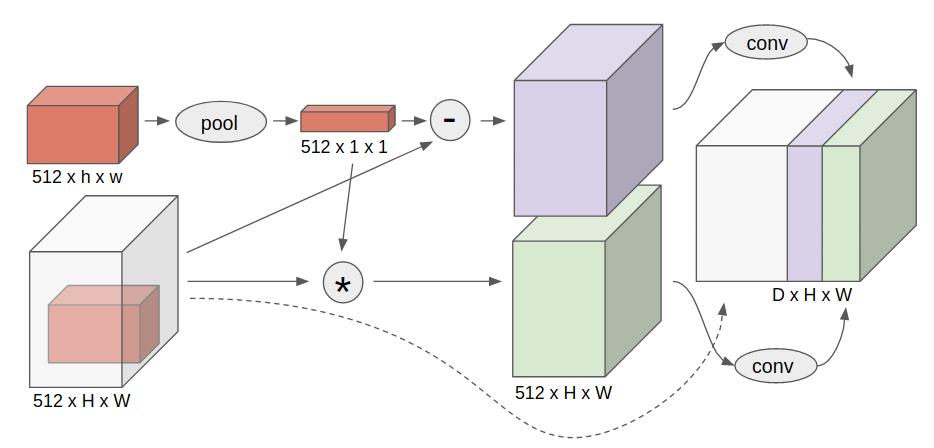}
\end{center}
   \caption{TDID embedding: given a pair of scene (gray) and target (red) features, makes a joint tensor embedding. Target features are pooled and then depth-wise correlated with (*) and subtracted from (-) scene features. In final model scene features (IMG in Table \ref{table:avd_ablation},  dotted line and white box in this figure) are not used. Features for one target image are shown, but multiple views of the same target may be added by concatenating their difference (green) and correlation (purple) features.  
}
\label{fig:tdid_embedding}
\end{figure}

\textbf{Joint Embedding} We construct a joint embedding, see Figure \ref{fig:tdid_embedding}, of all input images that can then be further processed for detection.  The joint embedding combines feature correlation and differencing between the target image(s) and the scene image. The operations and features in the embedding are described below and Table~\ref{table:avd_ablation} shows ablation results as different feature combinations are considered.

\textbf{Cross Correlation} is widely used in traditional methods with hand-crafted features for similarity matching. We started building the joint embedding by applying the cross correlation of target features with the scene features, generating a heatmap with only one channel dimension. This method generates a strong signal for predicting target presence/absence in each spatial location, but drops rich information from the feature channels. \emph{Depthwise-separable correlation} applies correlation at each channel independently. This not only preserves more information for the subsequent instance localization but also yields high computational efficiency~\cite{mobilenet,chollet2016xceptionDepthwiseSeparable}. We use depthwise-separable correlation in our joint embedding, represented as $CC$ in the ablation study, Table \ref{table:avd_ablation}, and the green box in Figure \ref{fig:tdid_embedding}.

\textbf{Feature differencing} is another way to measure similarity. Intuitively, a network attempting to learn a similarity between image features may do something like learn to subtract them. Instead of adding extra complexity to our framework by learning a similarity,  we compute the difference directly and feed it as a signal to our joint embedding. We first apply global max pooling on the target features to bring them to $1\times1$ spatial resolution. Then we subtract this vector from each spatial location of the scene features. This feature is represented as $DIFF$ in the ablation study, Table \ref{table:avd_ablation}, and the purple box in Figure \ref{fig:tdid_embedding}.

\textbf{Scene Image Features} The features of the scene image from the feature extractor may also provide useful information for object detection. In the original RPN of Faster-RCNN, these are the features that are used to predict bounding boxes and potential objects. This feature is represented as $IMG$ in the ablation study, Table \ref{table:avd_ablation}, and the white box in Figure \ref{fig:tdid_embedding}.

\begin{table}[!t]
\setlength{\tabcolsep}{0.17em}
\centering
\footnotesize{
    \begin{tabular}{|lcccc|c|}
         \hline
          Features Used & extra small & small &medium & large & All \\  
         \hline 
        \scriptsize{IMG}  & 1.9 & 7.7 & 5.1 & 5.3  & 2.2\\
        \scriptsize{CC}  & 23.8 & 58.5 &44.0 &50.7  & 27.7\\ 
        \scriptsize{DIFF}  & 48.0 & 74.6 & 72.3 & 73.2  & 52.6\\
        \scriptsize{IMG+CC}  & 28.0 & 54.5 & 51.4 & 54.8  & 31.9\\
        \scriptsize{IMG+DIFF}  & 46.2 & 79.2 &72.5 &71.3  & 50.9\\
        \scriptsize{CC+DIFF}  & \textbf{50.3} & 78.2 & \textbf{75.1} & \textbf{78.2}  & \textbf{55.8}\\
        \scriptsize{IMG + CC + DIFF} & 48.4 & \textbf{83.0} & 73.8  & 77.1 & 53.3\\
                 \hline
    \end{tabular}
    \caption{Ablation study of features in TDID embedding on various object sizes in AVD split 2 \cite{active-vision-dataset2017}. IMG==scene image features, CC==cross-correlation, DIFF==difference. mAP reported. }
\label{table:avd_ablation}
    }   
    \end{table}

\textbf{Ablation Study}
We run an ablation study to show how using different combinations of features in our joint embedding affects detection performance. Results are reported for the instance detection task on split 2 of AVD. As expected, using just scene image features, $IMG$, fails as there is no information about the target instance. Surprisingly, using just $DIFF$ features provides a strong signal resulting in high detection performance. The addition of $CC$ features provides a small boost in performance here, and also proved to be useful in later experiments so it is included in our final model. $IMG$ features do not provided much new information from $DIFF$ and $CC$, while adding extra complexity and parameters to the network and so are omitted from our final model.

\textbf{Final Embedding}
Our final joint embedding first pools the target features to be $N\times 1 \times 1$ where $N$ is the number of channels outputted by the feature extractor. This pooled target feature vector is then both cross-correlated with, and subtracted from, every location in the scene image feature map. These features, $CC$ and $DIFF$, are then each passed through their own $3 \times 3$ convolution to reduce the feature dimension to $\frac{N}{2}$. The $IMG$ features, represented by the dotted skip connection and white box in Figure ~\ref{fig:tdid_embedding}, are not used in the final model. The $CC$ and $DIFF$ features are then concatenated and passed through a final $3 \times 3$ convolution before being sent to the classification and regression filters. 

Figure ~\ref{fig:arch_full} shows the model for one target image and one scene image. In general, many target images may be used, providing more views of the target instance. Each target image will generate its own set of $CC$ and $DIFF$ features, which will all be concatenated before going through their respective $3 \times 3$ convolutions.

\textbf{Training}
To construct the training loss, we follow the region proposal settings in Faster R-CNN. For box localization regression we use Smooth $L_1$ error. Each anchor box is matched to the ground-truth target object box if its intersection-over-union (IoU)  is over $0.6$ and to background if its IoU is lower than $0.3$. Since we are only looking for one object at a time, there are only two classes for each box: target or background. 

\textbf{Inference}
During inference, we run one input/target pair at a time. In each case we select at most $5$ detections after non-maximum suppression with $0.7$ IoU threshold. We use IoU=$0.5$ as the matching criteria and modify the COCO evaluation parameters\footnote{https://github.com/cocodataset/cocoapi} for our experiments to report accurate mean Average Precision (mAP) results.

\section{Experiments}

We evaluate our method on three tasks: object instance detection, one-shot instance classification, and  few-shot object instance detection. For all TDID models we use Pytorch \cite{pytorch2017}, CUDA 8.0, and cuDNN v6.

\subsection{Object Instance Detection}
For all instance detection experiments, we report the same mAP as regular object detection. Since our system only considers one object at a time, to calculate mAP fairly we test all pairs of target object and scene image on our system. For example for AVD there are 30 instances. So for each image in the test set we run our network (or part of it, see below) 30 times, once for every instance. A general object detector runs once per image, and outputs boxes for every class. This seems like a big disadvantage for our system, since it is cumbersome to run the network for every single instance. In fact, this is where our system gains its advantage. In many applications, the system will only be looking for one, or very few, object(s) at a time. Our network is able to take advantage of this to greatly increase performance.

\begin{table}
\small
\parbox{\linewidth}{
\centering
\begin{tabular}{|cccc|}
   \hline
  Method & Backbone & image size & speed   \\
   \hline
  SSD\cite{liu15ssd} & VGG16&  512x512 & 19fps \\
  Faster-RCNN\cite{fasterRCNN} & VGG16&  600x1000 & 5fps \\
  TDID & VGG16&  960x540 & 12fps \\
  TDID & VGG16&  720x405 & 19fps \\
   \hline
   \hline
	\end{tabular}
\caption{Speed of various object detectors. Faster-RCNN\cite{fasterRCNN} and SSD\cite{liu15ssd} speeds are reported in their respective papers.  }
\label{table:Speed}
}
\end{table}

\begin{table}
\hfill
\parbox{\linewidth}{
\centering
\small
\begin{tabular}{|ccccc|}
   \hline
   \# of Instances & 1 & 2 & 5 & 10   \\
   \hline
  \shortstack{TDID  \\(960x540)} & 12fps&  10fps & 6fps& 4fps \\
  \shortstack{TDID  \\(720x405)} & 19fps&  16fps & 10fps& 6fps \\
   \hline
\end{tabular}
\caption{How the inference speed of TDID changes when detecting multiple instances in a single scene image, on a TITAN X GPU.}
\label{table:time_many_instances}
}

\end{table}

It should also be noted that \textbf{we do not need to run our entire system multiple times for multiple targets in one scene image}. Once the model is trained, all target features through the backbone feature extractor can be pre-computed and stored. Then features are extracted for each scene image once, and we only run the joint embedding and detection head of the network for each target. See Table ~\ref{table:time_many_instances} for a study of how inference time changes as more instances are detected in a single scene image. 

The speed/accuracy trade-off of object detectors has become of great interest in recent years\cite{speed-acc-tradeoff} as general object detectors get better and faster. Table \ref{table:Speed} compares the speed of TDID with the reported speeds of Faster-RCNN and SSD. TDID is a lightweight detector and can achieve speeds approaching that of SSD in settings where only a small number of objects are to be detected, while improving instance detection performance. It is well-known that use of various feature extraction backbone networks can influence detection performance and speed, and so we use VGG-16 in all experiments to keep comparisons fair.

\subsubsection{Active Vision Dataset}
\label{sec:active_vision_dataset}

We first evaluate our system on a challenging object instance detection dataset, AVD \cite{active-vision-dataset2017}. We use two target images (provided on the dataset website) for each instance, picking views to maximize how much of the object is seen. See Figure \ref{fig:target_input_ex} for an example of two target images. We choose two target images because in general it may be impossible to recognize an instance from the back if only the front view is provided.

We report results for all three train/test splits reported in \cite{active-vision-dataset2017}. For training, we resize all images to $960x540$ and use a learning rate of .001, momentum as .9 and weight decay 0.0005 and train for 40 epochs with . We then reduce the learning rate by a factor of 10, and continue training for
another 15 epochs. Table ~\ref{table:AVD2AVD} shows that our method outperforms SSD\cite{liu15ssd,active-vision-dataset2017}  on this task consistently, over 14 mAP on each split on all boxes (boxes $> 50 \times 30$). 

To produce a TDID system that runs at the same frame rate as SSD, we resize all images during testing to $720x405$. We test the same model that was trained on the $960x540$ images, and show results in the TDID(720x405) row in Table \ref{table:AVD2AVD}. We can see TDID still outperforms SSD by an average of over 5 mAp on all objects, and over 20 mAP on larger objects. We expect training a model at this resolution could result in even greater accuracy gains, while maintaining speed.

\begin{table}
	\begin{center}
    \begin{tabular}{|ccccc|}
   \hline
  Method & Box Size & Split 1 & Split 2 & Split 3  \\
   \hline
  SSD\cite{active-vision-dataset2017} &    \multirow{3}{*}{$> 100\times50 $} & 39 & 55& 53 \\
  TDID(720x405) & & 65.6 & 71.6&72.1 \\
  TDID(960x540)  & & \textbf{70.3} & \textbf{75.4}&\textbf{72.7} \\
   \hline
  SSD\cite{active-vision-dataset2017} & 
  \multirow{3}{*}{$> 50\times30 $} & 26 & 41& 42 \\
  TDID(720x405) &  &35.8 & 42.7& 48.2 \\
  TDID(960x540) &  & \textbf{48.9} & \textbf{55.8}& \textbf{56.5} \\
   \hline
	\end{tabular}
	\caption{Instance detection results (mAP) on the AVD dataset, with VGG16 backbone.}
    \label{table:AVD2AVD}
	\end{center}
\end{table}

\subsubsection{GMU Kitchens to AVD}

We now compare on a different object instance detection task to Faster-RCNN \cite{fasterRCNN}. \cite{cutpaste2017} explore how to create synthetic training data for instance detection, and evaluate how their synthetic data can improve a detector's performance when trained on one dataset, but tested on another. They train/test on the six instances present in both the GMU Kitchens dataset and AVD. In this task, the detector is trained on the GMU data, and tested on all images in the initial release of AVD (17,556 images). 

First, we train only on the real images from GMU, and test on AVD. We use the same training hyper-parameters as in the previous instance detection task. On this challenging task TDID is able to outperform Faster-RCNN by over 8 mAP.

Next, we add synthetic images to training. \cite{cutpaste2017} did not release their synthetic images, but did release code to generate them. We use their code and settings described in the paper to generate 5,160 synthetic images (\cite{cutpaste2017} report generating about 6000). Given extra training data, both Faster-RCNN and TDID improve. TDID retains its advantage over Faster-RCNN by 6 mAP, which may be further improved with better synthetic data.

\begin{table}
\begin{center}

\small
\parbox{\linewidth}{
\setlength{\tabcolsep}{0.17em}
\footnotesize{
    \begin{tabular}{|ccccccccc|}
        \hline
         Train set& Method & \shortstack{coca \\ cola} & \shortstack{honey \\ bunches} 
        & \shortstack{hunt's  \\sauce} & \shortstack{mahatma\\rice} & \shortstack{nature\\ v2}& \shortstack{red \\ bull}& mAP \\ 
        \hline
  \shortstack{Real \\Images}& \shortstack{Faster \\RCNN} & \textbf{57.7} & 34.4 & 48.0& 39.9&24.6  & 46.6 & 41.9  \\
  \hline
  \shortstack{Real \\Images}& \shortstack{TDID} & 57.4 & \textbf{34.5} & \textbf{73.8}& \textbf{43.3}&\textbf{32.1}  & \textbf{57.0} & \textbf{49.7} \\
  \hline
  \hline
  \shortstack{Real + \\Synthetic*}& \shortstack{Faster \\RCNN} & \textbf{69.9} & 44.2 & 51.0& 41.8& 48.7 & 50.9& 51.1  \\
  \hline
  \shortstack{Real + \\Synthetic}& \shortstack{TDID} & 69.1& \textbf{46.9} & \textbf{69.7} & \textbf{43.0}&\textbf{ 62.4}&\textbf{53.7}  & \textbf{57.5}  \\
                \hline
    \end{tabular}
    \caption{Detection performance (Average Precision) when training on GMU Kitchens and testing on AVD. *Synthetic images used in \cite{cutpaste2017} and ours are slightly different. }
    }   \label{table:GMU2AVD} }
\hfill
\end{center}
\end{table}

\begin{table}
\centering
    \begin{tabular}{|c|c|}
        \hline
         Method & Accuracy \\ 
         \hline
        Random & 0.3 \\
        BRISK \cite{BRISK} & 9.4 \\
        ORB \cite{ORB} & 6.6 \\
        SURF \cite{SURF} & 10.8 \\
        BOLD \cite{BOLD} & 7.4 \\
        SIFT \cite{SIFT} & 12.9 \\
        Line-2D \cite{LINE-2D} & .9 \\
        Color Hist \cite{color-hist-inter} & 9.2 \\
        HMP \cite{HMP} & 25.4 \\
        CaffeNet \cite{instanceHeld} & 41.0 \\
         CaffeNet+MV\cite{instanceHeld} & 44.1 \\
        \textbf{TDID(ours)} & \textbf{50.5} \\
  
    \hline
    \end{tabular}
    \caption{One-shot instance classification in a scene.}
\label{table:OneShotClass}
\end{table}

\subsection{One-Shot Instance Classification} \label{sec:classification}

We have shown our method outperforms state-of-the-art general object detectors on multiple instance detection tasks. We now show that we can also surpass other instance recognition and template matching work, as well as generalize to unseen target instances. \cite{instanceHeld} classify images of instances when given only a single image in training. They show a neural network, combined with some multi-view pre-training, can outperform previous non-deep-learning feature matching methods. \cite{instanceHeld} use a CaffeNet\cite{CaffeNet} classification network, pre-trained on ImageNet\cite{russakovsky2015imagenet}. They then perform a multi-view pre-training step on BigBIRD, train on a single example of each instance in the RGB-D Scenes\cite{UWASH-OBJECTS} dataset, and test classification accuracy on crops of instances in RBG-D Scenes.

We adapt our object detection framework to perform classification, and evaluate this modified network on the same one-shot instance classification task. In this setting, the definition of ``target image'' stays the same, but ``scene image'' is now a classification style image, i.e. a crop around one object. For TDID to generalize to unseen target instances it must be provided with a large variety of target instances during training. We construct a training set consisting of over 250 instances from the BigBIRD dataset and RGB-D Object Dataset, carefully excluding any instances that
overlap with those in the test set, RGB-D Scenes.

 For a fair comparison, we use AlexNet \cite{Krizhevsky_imagenetclassification} (extremely similar to CaffeNet\cite{CaffeNet}, same performance on ImageNet classification) pre-trained on ImageNET as our backbone network. To test how well our system can generalize to unseen target instances, we do not train on the single example of each test instance as \cite{instanceHeld} does. Instead we use the provided example as the target image at test time, \textbf{never re-training our network or updating the weights} to recognize these new objects. Even without any fine-tuning on the test objects our method achieves 50.5\% classification accuracy, outperforming the previous deep-learning approach that does train on the test objects, as well as several feature and template matching methods.  See Table~\ref{table:OneShotClass}.

\subsection{Few Shot Instance Detection}

\begin{table}
\setlength{\tabcolsep}{0.17em}
\centering
\footnotesize{
    \begin{tabular}{|c|c||c|}
        \hline
        object & TDID & Faster-RCNN\cite{cutpaste2017} \\
        \hline
        coca cola &30.8  &88.5*\\
        coffe mate  & 73.8 & 95.5*  \\
        honey bunches& 52.0 & 94.1* \\
        hunt's  sauce & 24.1 & 88.1* \\
        mahatma rice& 26.7 & 90.3* \\
        nature v1   & 86.1  & 97.2* \\
        nature v2  & 82.2  & 91.8*  \\
        palmolive orange  &28.3  &80.1* \\
        pop secret  &62.2  &94.0* \\
        pringles bbq   &26.0  &92.2* \\
        red bull   &37.9  &65.4* \\
        \hline 
        mAP    &48.2  &88.8* \\
 
                \hline
    \end{tabular}
    \caption{Few-shot detection average precision on GMU Kitchens. Instances were not seen as targets during training for TDID, though nature v1/v2 are similar to training instances. *Just a reference, as Faster-RCNN trains on these instances.}
\label{table:GEN4GMU}
    }   
\end{table}

We next explore few-shot instance detection with two examples of each instance available, one front view and one back view. In contrast with usual few-shot tasks, \emph{we do not train on examples of test objects}. We use the examples as target images at test time, requiring our detector to generalize to unseen objects without any on-line training or fine-tuning. High performance on this task could be useful for many applications where the system is given just a few examples of a target object but does not have time to re-train. We test on the instances in the GMU Kitchens dataset.

 
 As in the previous experiment, to enable TDID to generalize we constructed a training set with many different target instances, from AVD, BigBIRD, RGB-D Objects, RGB-D Scenes, and ImageNET VID. VID consists of snippets of video with one or more objects labeled with a bounding box. While training TDID, we first choose a video at random, then choose an object as the target, and crop two random frames of the video to get target images. Another frame from the video is chosen as the scene image 50\% of the time, while a frame from a different video is chosen the other 50\%. This means the target object is visible in the scene image in half of the examples. 
 We use the same instances from BigBIRD and RGB-D Objects as in Section \ref{sec:classification}, but instead of cropped classification images we use code from \cite{cutpaste2017} to synthetically place the objects in 1449 images from NYUD2\cite{nyud2}. 
 
  In addition to the released bounding boxes in AVD, we take advantage of the dataset's structure to add more target instances automatically. Starting with an image, $I$, in a scene, $S$, we use selective search to get the bounding box of some object or region, $O$, in $I$. Using the camera locations and depth images provided by AVD, we can project $0$ to world coordinates and then project back into every other image in $S$ to get the bounding box of $O$ in every image. This gives us more target instances almost for free. Unfortunately this setup is still experimental, and is not always robust to occlusion and other factors. Therefore we only generate samples from two scenes from AVD, adding ~5000 target/scene image pairs. Future work includes making this process more robust to hopefully greatly increase TDID's generalizability. 

We use the same training hyper-parameters as in the detection experiment on AVD in Section \ref{sec:classification}, except we cut the learning rate in half to $.0005$ and train for 150,000 iterations. 


 As shown in Table ~\ref{table:GEN4GMU}, TDID is able to generalize well to these instances, achieving 48.2 mAP. We also provide the Faster-RCNN results \cite{cutpaste2017} from training/testing on split one of the GMU data as a sort of upper bound reference, and to show the difficulty of the GMU data relative to other tasks. This result is particularly exciting as TDID is able to give reasonable performance on a task general category detectors cannot perform. The ability to detect novel objects quickly, without any new training, could be very valuable for robots in many applications.

\section{Analysis, Limitations, and Future Work}

TDID is able to outperform previous object recognition systems at a variety of challenging tasks. Given the comparison of a target to a potential detection, the remaining classification problem is binary, matching the target or not. Category detection methods need to discriminate between many different classes, both requiring more training data per class, and often a second stage network (e.g. in the Faster RCNN, consisting of RPN followed by classification and final bounding box regression). Our method is based on the idea that it is sometimes easier to learn to compare two things, than to learn about every object. This may be especially relevant when a small number or only one training example is used as shown in the results in Table 6 where TDID outperforms a wide range of methods on single shot classification.

Our method still has the same difficulty of detecting small objects as other detectors, as can be seen in the drop in performance as box size changes in Table \ref{table:AVD2AVD}. In addition, while TDID's ability to detect objects it has never seen during training (Table \ref{table:GEN4GMU}) is exciting, the objects it can generalize to are still limited. All of the train/test objects came from a similar household/grocery store domain.

Future work includes improving the generalizability and detection performance of TDID to objects across many different domains. We hope in the future an object detection system will be able to work off the shelf, detecting any objects a robot or other system may be interested in without the need for training. 






\bibliographystyle{IEEEtran}
\bibliography{references}

\end{document}